\documentclass[conference]{IEEEtran}
\IEEEoverridecommandlockouts
\usepackage{cite}
\usepackage[utf8]{inputenc}
\usepackage{graphicx}
\usepackage{amsmath}
\usepackage{amssymb}
\usepackage[version=4]{mhchem}
\usepackage{siunitx}
\usepackage{longtable,tabularx}
\usepackage{subcaption}
\usepackage{epstopdf}
\usepackage{tabularx}
\usepackage{booktabs} 
\usepackage{multirow}
\usepackage{arydshln} 
\usepackage{url}
\usepackage{float}
\usepackage{acronym}
\usepackage{fancyhdr}
\usepackage{threeparttable}
\usepackage{comment}
\usepackage{hyperref} 
\usepackage{textcomp} 
\usepackage{makecell}
\DeclareMathAlphabet{\mathcal}{OMS}{cmsy}{m}{n}

\usepackage{hyperref} 
\usepackage{textcomp} 
\def\BibTeX{{\rm B\kern-.05em{\sc i\kern-.025em b}\kern-.08em
    T\kern-.1667em\lower.7ex\hbox{E}\kern-.125emX}}

\renewcommand{\footnoterule}{%
  \kern -3pt
  \hrule width \columnwidth height 1pt
  \kern 2pt
}

\begin{document}

\title{Learning-based Directed Graph Abstraction of Combinatorial Spaces for Order-Preserving Search in Mixed-Combinatorial Nonlinear Optimization}

\makeatletter
\newcommand{\linebreakand}{%
  \end{@IEEEauthorhalign}
  \hfill\mbox{}\par
  \mbox{}\hfill\begin{@IEEEauthorhalign}
}
\makeatother

\author{
\IEEEauthorblockN{Gishnu Madhu$^*$\thanks{$^*$ M.S. Student, Department of Computer Science and Engineering}}
\IEEEauthorblockA{\textit{University at Buffalo}\\
Buffalo, New York, 14260 \\
gishnuma@buffalo.edu}
\and
\IEEEauthorblockN{Feng Liu$^\dagger$ \thanks{$^\dagger$ Ph.D. Candidate, Department of Mechanical and Aerospace Engineering}}
\IEEEauthorblockA{\textit{University at Buffalo}\\
Buffalo, New York, 14260 \\
fliu23@buffalo.edu}
\and
\IEEEauthorblockN{Souma Chowdhury$^\ddag$ \thanks{$^\ddag$ Professor, Department of Mechanical and Aerospace Engineering; Professor (adjunct), Department of Computer Science and Engineering, Corr. author} \thanks{This work is accepted to be presented at the 2026 IDETC-CIE.}
}
\IEEEauthorblockA{\textit{University at Buffalo}\\
Buffalo, New York, 14260 \\
soumacho@buffalo.edu}
}

\maketitle
\thispagestyle{plain}
\pagestyle{plain}


\newacro{ADR}{Active Debris Removal}
\newacro{RL}{Reinforcement Learning}
\newacro{CO}{Combinatorial Optimization}
\newacro{MINLP}{Mixed Integer Non-Linear Programming}
\newacro{MCNLP}{Mixed Combinatorial Non-Linear Programming}
\newacro{MDPSO}{Mixed-Discrete Particle Swarm Optimization}
\newacro{GNN}{Graph Neural Network}
\newacro{GA}{genetic algorithm}
\newacro{COMPASS}{Combinatorial Optimization with Policy Adaptation using Latent Space Search}
\newacro{COMBO}{Combinatorial Bayesian Optimization}
\newacro{PBGE}{Preference-Based Gradient Estimation}
\newacro{GFACS}{Generative Flow Ant Colony Sampler}
\newacro{GFlowNets}{Generative Flow Networks}
\newacro{MAXQ}{MAXQ Value Function Decomposition}
\newacro{UAV}{Unmanned Aerial Vehicle}
\newacro{MLP}{multilayer perceptron}

\begin{abstract}
Mixed-combinatorial nonlinear programming (MCNLP) problems are a challenging class of optimization problems that arise in various engineering design and optimal planning applications, e.g., due to the combination of categorical, component, and geometric choices in systems design, and the need for joint task and motion planning. Traditional approaches to represent the combinatorial space in conjunction with the continuous space include integer indexification or binary representations, which, however, introduce spurious (unintended) relations between combinations, lead to a substantial increase in the search space dimensionality, and/or require imposing additional compatibility constraints. Instead, this paper draws on recent developments in robot planning and vehicle/network routing domains that aim to learn search heuristics over combinatorial spaces using graph neural networks (GNNs). More specifically, this paper presents a first-of-its-kind structured abstraction of the combinatorial space by learning to map a set of combinations (combinatorial choices) represented as an undirected, fully connected graph to a directed graph where edges indicate the direction of improvement. The mapping is performed by an Edge Field Graph Network (EFGN) trained via pairwise difference regression, with a loss function that uses a special cyclical regularizer. To demonstrate the utility of this new way of abstracting the combinatorial space in solving MCNLPs, we adopt a recent optimization framework that purely searches over the non-combinatorial (e.g., continuous) variables and retrieves the best-suited combination for each candidate design by using the abstraction model (EFGN), akin to a recommender system. The presented direction-aware abstraction model provides a potentially more scalable and interpretable retrieval of combinations compared to the original recommendation system in that framework. For evaluation purposes, the optimization is implemented using well-known particle swarm optimization and genetic algorithm solvers for three benchmark nonlinear (constrained and unconstrained) problems that vary in the number of combinations and variables. Compared to baseline optimization solvers using indexification of the allowed combinations, the optimization using the GNN-based recommender consistently provides a much better mean optimum value and robustness across multiple runs of the solver in each case. 
\end{abstract}




\section{Introduction}
Many complex real-world engineering and scientific problems are characterized by the simultaneous optimization of discrete choices along with continuous parameters. Examples of such types of problems commonly exist in the field of robot design~\cite{robotdesign}, spacecraft design~\cite{gnn_reco, jsr2023Zeng}, and task planning~\cite{Nikitina04042025}. A more representative example can be found in \ac{UAV} design~\cite{uav_zeng}, where certain variables, such as propeller length, are continuous, while others—such as motor size—can be discrete and predefined by manufacturers. Moreover, these discrete components are often only compatible with specific subsystems (e.g., batteries), leading to constrained combinations that must be jointly considered during the optimization process. The presence of continuous variables distinguishes these problems from classical \ac{CO} problems such as network management~\cite{GraphColorFlowManagement}, the traveling salesman problem~\cite{ReviewofCo}, and graph coloring~\cite{GraphColorFlowManagement}, which typically involve only integer or combinatorial decision variables and operate over a finite discrete search space. Not to mention, they can be formulated as linear programming problems, which is unviable for most complex system design problems.
In this paper, we focus on optimization problems characterized by the co-presence of continuous and combinatorial variables, which are broadly categorized as Mixed Combinatorial Nonlinear Programming (MCNLP) problems. Our objective is to develop an efficient optimization framework that jointly optimizes both continuous and combinatorial variables. Building on our previous work~\cite{gnn_reco}, we propose to employ a \ac{GNN} to recommend the best-suited combinatorial variable vector conditioned on a given continuous variable vector.

In \ac{MCNLP} engineering design problems, decision variables may include continuous parameters (e.g., geometry), integer variables (e.g., counts of features such as fins or holes), and selections from finite sets of combinations. Such combinations typically arise from component or material choices, where each option is characterized by multiple attributes, such as motor mass, voltage, and power ratings, or material density and elasticity. When these choices are independent, they can be treated as separate combinatorial sets, whereas dependencies or conflicts among components require defining a unified set of valid combinations. Unlike continuous or integer variables, combinations do not naturally support vector operations or ordered neighborhood relationships, which makes them difficult to handle using conventional gradient-based or metaheuristic optimization methods. Consequently, most existing approaches treat the combinatorial space as a selection problem or transform it into alternative representations that are more amenable to optimization, such as one-hot or ordinal encoding with integers~\cite{LiuZhihao2023BOoM, XIA2025103415, OkadaShuntaro2019Epoi}.

These problems are computationally NP-Hard \cite{Relaxation}, and the increased choices among discrete variables lead to combinatorial explosion\cite{MINLP-NPHard}. Therefore, the mixed combinatorial search space is generally intractable, which necessitates abstractions to represent it. When solving optimization problems, this is typically done via subsets, projection, or latent representations as abstractions. For example, Pattern Databases, which is a hierarchical method, abstracts combinatorial spaces by reducing them to simplified subproblems, providing tight lower bounds that guide and accelerate search\cite{patterndatabase}. \ac{COMPASS} abstracts combinatorial space by mapping it to a low-dimensional continuous latent representation of policies, within which effective strategies are searched\cite{Compass}. \ac{MAXQ} abstracts by using only subtask-relevant variables under safety guarantees, reducing the complexity \cite{MAXQ}.
\ac{COMBO} on the other hand, abstracts the combinatorial space by representing combinations as nodes in a graph and models a surrogate function over this using Laplacian-based kernels, exploiting local transitions between combinations\cite{COMBO}. While these methods expose techniques to reduce the complexity of the search space, they lack a structured notion of directionality within the combinatorial space. They are thus limited in their ability to provide navigational guidance during optimization. 

Several existing research studies aim to improve navigation or provide gradient-like guidance for effective search in the combinatorial space. Gumbel-Softmax relaxation~\cite{gumbelSoftmax} technique brings gradient awareness to combinatorial exploration, but it suffers from a discretization gap and scalability bottlenecks. \ac{GFACS} is a technique were combinatorial space is abstracted as a sequential decision process over a directed acyclic graph (DAG), here a \ac{GFlowNets} learns a reward distribution over solutions using \ac{GNN} internally. This learned prior further guides ant-based search toward high-quality solutions~\cite{antcolony}. Despite the significant progress, the existing methods operate at the level of state compression or scalar function approximation. Most existing methods assume a static combinatorial objective, whereas in MCNLP settings the landscape itself varies with the continuous context, requiring a representation that adapts across conditions. These limitations point to a critical gap: there is a lack of a unified, geometry-aware abstraction and learning methodology that can learn transitions in the objective landscape, especially in the mixed combinatorial domain. 

A recent approach for solving such \ac{MCNLP} problems in design automation presents a decomposition concept of separating the \ac{CO} part from the problem and solving the \ac{CO} problem with the aid of \ac{GNN} as a recommender for combination selection. This concept partly draws motivation from work in multi-robot task allocation~\cite{paul2024learning, ref3} and network re-configuration~\cite{ref4}, where large combinatorial spaces are effectively represented as graphs and efficiently searched over by GNNs trained on data or experience. Building on this concept, our recent work~\cite{gnn_reco} presents \textbf{GNN-ReCo}, an optimization framework that divides the problem into a combination learning stage—where combinations are represented as graph nodes and processed by a GNN-based recommender—and a non-\ac{CO} optimization stage, which is solved using standard NLP or \ac{MINLP} methods.

Liu et al.~\cite{gnn_reco} abstract the combinatorial search space as a graph and propose a \ac{GNN}-based recommendation system, named as \textbf{GNN-ReCo}, to embed in the unconstrained optimization framework, which efficiently selects combinatorial variables corresponding to a given continuous variable vector. Building on this optimization framework, we further draw inspiration from abstracting combinatorial spaces through directed graphs of pairwise relationships, where global structure is inferred from local edge-wise comparisons. This perspective \cite{HodgeRanks} serves as our novel formulation for building a learning-based model to reason over combinatorial structures in a geometry-centric way and to learn edge-wise objective differences between combinations indexed by continuous variables. We extend the GNN-ReCo Optimization framework and learning edge-wise energy differences to a mixed combinatorial optimization setting.

In this paper, we explore a different approach for the recommender system, which could potentially also enhance the scalability and interpretability of the search process. The main hypothesis behind this new approach is this notion of explicit gradient awareness over combinatorial spaces -- we posit that by learning explicit quantification of signed change in criteria functions when moving from one combination to another over large combinatorial spaces, we can make the retrieval more scalable and the recommendation process more structured and interpretable.
This new approach necessitates a new graph representation, where edge weights encode relative objective differences between combinations. Through this construction, we define a structured mathematical object over the combinatorial graph that captures the local geometry of the optimization landscape. In practice, during each combinatorial choice selection step, the model is provided with a subset of candidate combinations, represented as nodes in an undirected graph, along with a continuous context proposed by the optimizer. The model then outputs a directed graph over the same nodes, where edges are assigned normalized objective differences. An illustrative visualization of this mapping concept is shown in Fig.~\ref{fig:gnn_navco}. This enriched directed graph then enables the optimizer to make an informed decision over the objective landscape, which is a mixed combinatorial hybrid search space. Consequently, this technique enables gradient-like navigation over this hybrid space while preserving the discrete structure and underlying geometry. This new GNN structure navigates the selection of the combinations; we name this variation of the prior recommending system as \textbf{GNN-NavCo} to distinguish from the prior work.

The key contributions of the paper are as follows:\\
\textbf{1) \textit{Abstraction of Combinatorial Spaces via Pairwise Energy Differences:}} We represent the searchable combinatorial space as a directed graph, where edges encode \emph{pairwise energy differences} between combinations. These values capture how the objective and penalty-augmented constraints evolve across transitions. We further formulate a mathematical framework over this representation, inducing a directional structure that provides gradient-like information to guide the search toward promising nodes.\\ 
\textbf{2) \textit{Gradient-aware Learning of Combinatorial Navigation via GNN:}} We develop a learning framework called \textbf{GNN-NavCo} that computes the gradient-like edge weights of the graph and predicts the corresponding \textit{energy} landscape for a given continuous variable vector of a candidate design. The predicted landscape is then used to guide the recommendation of promising nodes, enabling efficient exploration of the combinatorial space across different sampled subsets of combinations.\\ 
\textbf{3) \textit{Evaluation of GNN-NavCo-aided optimization:}} We decompose the \ac{MCNLP} solution process into a combination recommendation model generation stage and a non-combinatorial (in our problems, continuous) optimization stage as \cite{gnn_reco}. We then conduct an extensive evaluation of the proposed framework on a set of two unconstrained benchmark problems to examine its scalability and performance across large combinatorial spaces. Furthermore, the framework is evaluated on constrained benchmark problems to demonstrate its applicability to constrained optimization settings. Standard particle swarm optimization (PSO) and genetic algorithm (GA) implementations are used to perform the optimizations demonstrating cross-algorithm applicability of the GNN-NavCo model. 


\begin{figure*}[t] 
    \centering
    \includegraphics[width=0.9\textwidth]{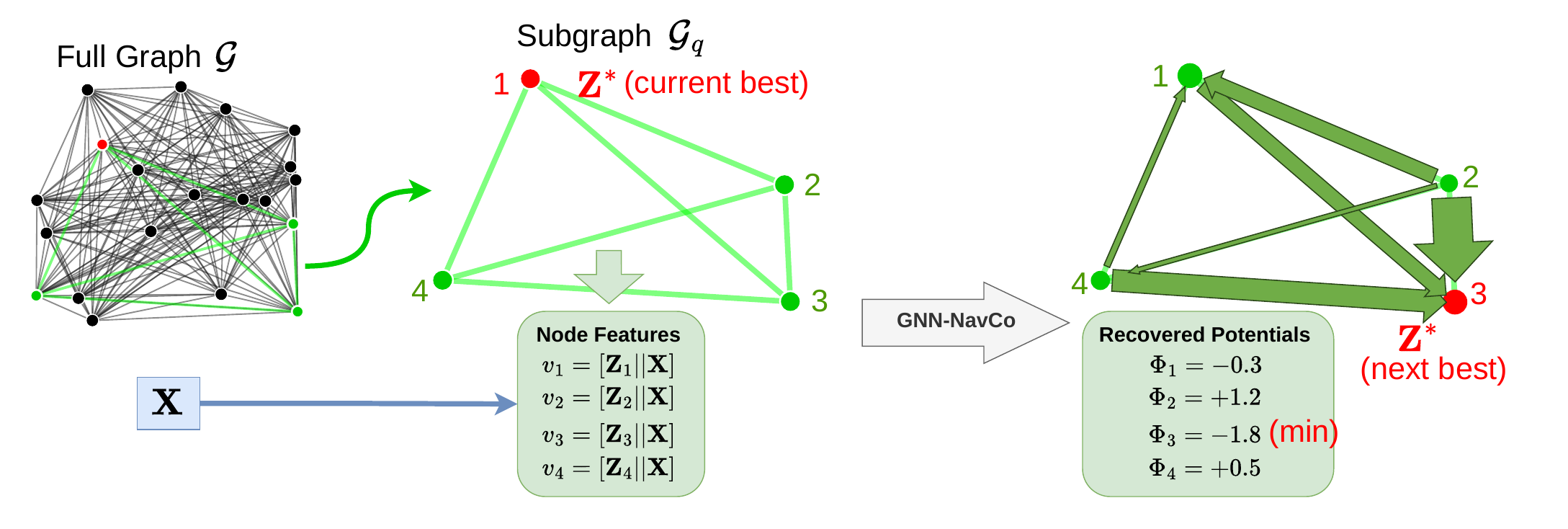}
    \caption{Example of GNN NavCo Inference Process: Takes as input a subset of allowed combinations expressed as a graph along with a candidate design vector, and produces as output a directed graph encoded with improvement direction and potential across the combinations.}
    \label{fig:gnn_navco}
\end{figure*}

\section{Mixed-Combinatorial Optimization Problem Formulation and Abstraction}\label{sec:mco}

\subsection{Mixed Combinatorial Optimization Problem}
We consider a \ac{MCNLP} with $k_{\text{comb}}$ combinatorial variables $z$ and $k_{\text{cont}}$ continuous variables $x$ with $p$ inequality constraints and $q$ equality constraints in the following form:
\begin{equation}
\label{eq:obj_raw}
\begin{aligned}
 \min_{\mathbf{Z}_{\text{comb}}, \mathbf{X}_{\text{cont}}} &f( \mathbf{Z}_{\text{comb}}, \mathbf{X}_{\text{cont}} ) \\
\text{subject to} \quad & g_i(\mathbf{Z}_{\text{comb}}, \mathbf{X}_{\text{cont}}) \leq 0, \quad i = 1, 2, \dots, p \\
& h_j(\mathbf{Z}_{\text{comb}}, \mathbf{X}_{\text{cont}}) = 0, \quad j = 1, 2, \dots, q \\
\text{where} \quad &  \mathbf{Z}_{\text{comb}} = \begin{bmatrix} 
z_1 & z_2 & \dots & z_{k_{\text{comb}}}\end{bmatrix} \\
&  \mathbf{X}_{\text{cont}} =\begin{bmatrix} x_1 & x_2 & \dots & x_{k_{\text{cont}}}
\end{bmatrix}\\
& x_n^L \leq x_n \leq x_n^U, \quad \forall ~n=1, 2, \dots, k_{\text{cont}}\\
& \mathbf{Z}_{\text{comb}} \in \mathbb{Z}
\end{aligned}
\end{equation}
Where $\mathbf{Z}_{\text{comb}}$ and $ \mathbf{X}_{\text{cont}}$ are the vectors of the combinatorial and continuous variables, respectively. $x^{\text{L}}$ and $x^{\text{U}}$ are the lower and upper bounds of the continuous variables, and $\mathbb{Z}$ is the valid combination set. With the aid of \textbf{GNN-NavCo}, the problem is decomposed and rewritten in the following Non-\ac{CO} form as in~\cite{gnn_reco}:
\begin{equation}
\label{eq:obj_change}
\begin{aligned}
 \min_{\mathbf{X}_{\text{cont}}} &~f( \mathbf{Z}^*_{i}, \mathbf{X}_{\text{cont}} ) \\
\text{where} \quad 
& \mathbf{y} = {\gamma}(\mathcal{G}, \mathbf{X}_{\text{cont}}) \\
& i^* = \underset{i=1, 2, \dots, n_{\text{tot}}}{\text{argmax}} y_i, \quad y_i \in \mathbf{y} \\
&  \mathbf{X}_{\text{cont}} =\begin{bmatrix} x_1 & x_2 & \dots & x_{k_{\text{cont}}}
\end{bmatrix}\\
& x_n^L \leq x_n \leq x_n^U, \quad \forall ~n=1, 2, \dots, k_{\text{cont}}\\
& \mathbf{Z}^*_{i} \in \mathcal{G},~{\text{and}}  ~\mathcal{G}\in\mathbb{Z}
\end{aligned}
\end{equation}
Here $\mathbf{y}$ is the score vector indicating the fitness value of each candidate combination in the abstracted graph $\mathcal{G}$, which is calculated by a function $\gamma(\cdot)$ involving the prediction of \textbf{GNN-NavCo}. $i^*$ is the index of the node (candidate combination in $\mathcal{G}$) with the highest fitness score, denoted by $\mathbf{Z}^*_{i}$.

A common approach in constrained optimization is to convert the constrained problem into an unconstrained formulation by introducing penalty functions that incorporate constraint violations into the objective~\cite{gnn_reco}.
Following this approach, we define a penalized energy function. We model the objective function landscape as an energy function~\cite{LeCun2006ATO}, where both feasible and infeasible objective values are unified into a single objective or energy landscape, yielding a mixed combinatorial energy landscape on which optimization is performed. The energy landscape is shaped by the objective function and constraints in equation~\eqref{eq:obj_raw}. 
\begin{equation}
\label{energy}
\begin{aligned}
f_E(\mathbf{Z},\mathbf{X}) = f(\mathbf{Z},\mathbf{X}) + \rho \Big(
& \sum_{i=1}^{p} P_\tau(g_i(\mathbf{Z},\mathbf{X})) \\
& + \sum_{j=1}^{q} P_\tau(h_j(\mathbf{Z},\mathbf{X})) \Big)
\end{aligned}
\end{equation}
where $\rho>0$ is a penalty weight and $P_\tau(\cdot)$ denotes penalty. It can be chosen based on the objective landscape. 

In the case of a minimization problem, the resulting optimization problem becomes $\min_{\mathbf{Z}, \mathbf{X}} f_E(\mathbf{Z},\mathbf{X})$. For a fixed continuous context $\mathbf{X}$, this reduces to $\min_{\mathbf{Z}} f_E(\mathbf{Z},\mathbf{X})$, which defines an energy landscape over the combinatorial space. In the following sections, we represent this landscape through differential relations between combinations using a graph-based formulation.


\subsection{Prior Work of GNN-ReCo}
The prior work of \cite{gnn_reco} decomposes the \ac{MCNLP} into two parts: the learning of combinations which are abstracted as graph nodes, and solving a Non-\ac{CO} as just NLP or \ac{MINLP}. If the valid combination set is defined as $\mathbb{Z}$, then each combination can be represented by a feature vector $\mathbf{Z} = [z_1, z_2, \dots, z_L]$, where $\mathbf{Z} \in \mathbb{Z}$ and $L$ denotes the number of features. These features may correspond to the physical attributes associated with the selected components. For example, in robot design, such features may include motor mass, voltage, and power ratings, as well as sensor detection range and mass. The work of \cite{gnn_reco} represents the combinatorial space as a fully connected undirected graph $\mathcal{G} = \{\mathcal{Z}, \mathcal{E}\}$, where $\mathcal{Z}$ is the node feature and $\mathcal{E}$ is the edge feature. Each row in the matrix $\mathcal{Z}$ is a valid combination $\mathbf{Z}$ that is abstracted as a node positioned according to its feature vector value. $\mathcal{E}$ is defined as the adjacency matrix of the Euclidean distance between each node. Trained by a specialized list-wise loss function, \textbf{GNN-ReCo} efficiently learns to predict the \textit{score} of each combination and then recommend the best-suited combination in the valid set for the given non-combinatorial variable vector $\mathbf{X}$. Here, the \textit{score} can be a scalar value defined by the objective function. By embedding it into the function evaluation, \textbf{GNN-ReCo} can work with both sampling-based heuristic optimization algorithms and classic optimization solvers. Additionally, the work of \cite{gnn_reco} proposed to train with subgraphs $\mathcal{G}_q$ (e.g., a sampled subset of the entire valid set) instead of the full graph $\mathcal{G}$. This benefits the \ac{GNN} majorly in two aspects: \textbf{1)} The training is more efficient and less expensive. \textbf{2)} If more valid combinations are added to the problem, the \ac{GNN} still maintains a level of robustness and can also be fine-tuned with the new combination sets without retraining from the beginning.

\subsection{Graph Abstraction of Energy Landscape}\label{ssec:landscape}

In this paper, modifications to the graph abstraction focus primarily on the edge feature $\mathcal{E}$ utilizing the energy function $f_E$ defined in Eq.~\eqref{energy}. Similarly, the proposed framework will also work on the sampled subgraphs instead of on the full graph.


In the original work of \cite{gnn_reco}, the topology of the fully-connected, undirected graph is independent from the continuous variable vector $\mathbf{X}$, and each node is associated with an energy value $f_E(\mathbf{Z}, \mathbf{X})$, inducing an energy landscape. Starting from such a graph, this paper introduces the pair-wise comparison based on $f_E(\mathbf{Z}, \mathbf{X})$ across the combinatorial space into the topology of the graph, specifically on the graph edges. For each pair of combinations $(\mathbf{Z}_i,\mathbf{Z}_j)$, we define an edge function $\omega(i,j,\mathbf{X})$, which captures the relative change in energy between combinations under context $\mathbf{X}$.

\begin{equation}
\label{omega}
     \omega(i,j,\mathbf{X}) = f_E(\mathbf{Z}_j,\mathbf{X}) - f_E(\mathbf{Z}_i,\mathbf{X})
\end{equation}

This construction induces a directed edge structure over the graph, where edge weights encode transition-wise energy differences. Importantly, while the graph topology remains fixed, the edge weights vary with the continuous context, thereby altering the relative ordering and direction of the improvement between combination choices. Furthermore, this edge quantity satisfies the antisymmetry property, 
$\omega(i,j,\mathbf{X}) = -\omega(j,i,\mathbf{X})$,
reflecting that reversing a transition negates the corresponding energy difference. Since the graph topology is fixed, we denote the context-conditioned function as
$\omega_{\mathbf{X}}(i, j) := \omega(i, j, \mathbf{X})$
representing the energy difference between combinations under a given context \( \mathbf{X} \). From this perspective, the continuous variables $\mathbf{X}$ parameterize a family of energy landscapes over a fixed combinatorial graph. Learning the edge function $\omega_{\mathbf{X}}(i, j)$ for each of the ordered combination pairs, therefore, enables the model to predict the context-dependent adjacency structure or edge weights of the graph. This allows the model to transform an undirected graph into a directed graph of combinations, as shown in Fig.~\ref{fig:gnn_navco}, enabling direction-aware navigation over the combinatorial space without altering its underlying structure.

This abstraction points to a structured object over the graph, which we refer to as an \emph{edge field}, that encodes differences in energy between combinations. This viewpoint admits a natural geometric interpretation: in the language of discrete differential geometry, scalar functions defined on vertices correspond to discrete 0-forms. In contrast, functions defined on edges correspond to discrete 1-forms \cite{crane2018ddg,desbrun2005discreteexteriorcalculus}. A discrete gradient is a special case of a 1-form arising from differences of a scalar potential. In this sense, $\omega_{\mathbf{X}}(i,j)$ can be interpreted as encoding magnitude and directional information of potential improvement over the mixed combinatorial space.




Thus, \(\omega_{\mathbf{X}}(i,j)\) defines a scalar quantity on the edges of the graph, capturing the change in energy between combinations and thus captures the gradient signal between them. In the case where this edge field corresponds to an exact form, i.e., it arises from an underlying scalar energy function, it must satisfy a fundamental consistency condition: the net change in energy along any closed cycle should be zero\cite{hirani2011squaresrankinggraphs}\cite{crane2018ddg}. Formally, for any cycle \(C\) in the graph,

\begin{equation*}
\sum_{(i,j)\in C} \omega_{\mathbf{X}}(i,j) = 0
\end{equation*}

In the complete graphs, triangular cycles form the simplest such constraints, and for any triplet \((i,j,k)\),
\begin{equation}
\omega_{\mathbf{X}}(i,j) + \omega_{\mathbf{X}}(j,k) + \omega_{\mathbf{X}}(k,i) = 0
\notag
\end{equation}

This condition is both necessary and sufficient for the existence of a globally consistent scalar energy function whose pairwise differences recover \(\omega_{\mathbf{X}}(i,j)\). In practice, however, the learned edge values may violate this property due to approximation errors, leading to non-zero cycle sums. While a consistent ranking can still be recovered via least-squares potential recovery \cite{hirani2011squaresrankinggraphs}, large violations introduce inconsistencies in the reconstructed energy landscape.

\section{LEARNING EDGE FIELDS ON THE GRAPH}

Given the graph $\mathcal{G}$ constructed from sampled combinations, we introduce an Edge Field Graph Network (EFGN) Architecture, a model that learns a parametric approximation of the differential edge field introduced in Section~\ref{sec:mco}. 
For a given context $\mathbf{X}$, the model predicts $\hat{\omega}_\theta(i,j;\mathbf{X})$ as an approximation of the true energy differences Eq.~\eqref{omega}. The proposed model adopts an encoder--decoder architecture. A \ac{GNN} encoder first processes the graph to produce context-conditioned node embeddings that capture the relational structure among combinations. These embeddings are then passed to a pairwise edge decoder, implemented as a \ac{MLP}, which maps pairs of node embeddings to predicted energy differentials. This design enables the model to learn a context-dependent differential representation of the energy landscape, where node-level representations encode global structure, and the decoder recovers directional relationships between combinations. The following subsections describe the encoder and decoder components in detail.

\begin{figure}
    \centering
    \includegraphics[width=0.8\linewidth]{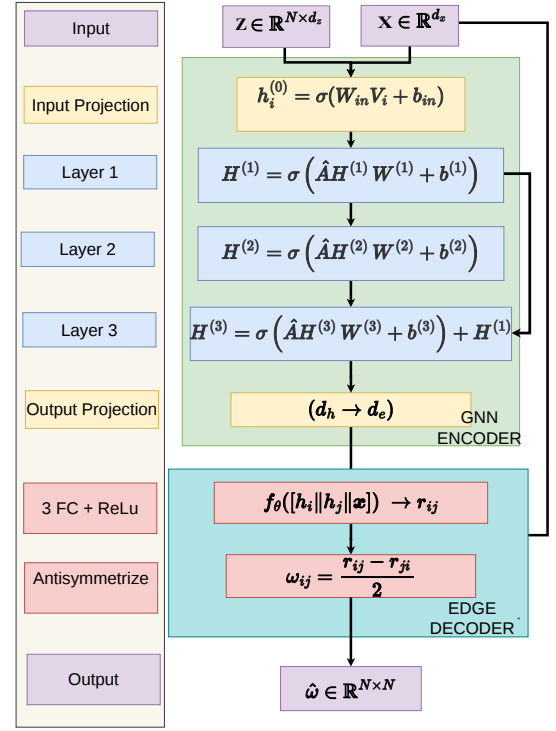}
    \caption{EFGN Model Architecture. Here $N$ represents the number of nodes in the graph. $d_z$ and $d_x$ are the number of combinatorial and continous variables, respectively.}
    \label{fig:arc}
\end{figure}
\subsection{GNN Encoder}

Before the raw context is passed onto the GNN, each combination is augmented with a shared continuous context vector. This ensures that all node representations are explicitly conditioned on the continuous variables. This step creates the $v_i = [\mathbf{Z}_i \parallel \mathbf{X}]$ for each combination. These concatenated features are first projected into a latent space $h_i^{(0)} = \sigma(W_{in} v_i + b_{in})$,  where $W_{in}$ is a learnable projection matrix and $\sigma(\cdot)$ denotes a nonlinear activation function and $b_{in}$ is the bias vector. A stack of graph message passing is applied, which allows each node to aggregate information from all other combinations, encoding its relative position within the candidate set. This can be expressed as : 
\begin{equation*}
    H^{(k+1)} = \sigma\left(\hat{A} H^{(k)} \, W^{(k)} + b^{(k)}\right) + H^{(k)}
\end{equation*}

where $H^{(k+1)} $ is node embeddings at layer k+1 , $\hat{A}$ is the row normalized adjacency matrix and $+H^{(k)}$ denotes the residual connection.  After $L$ layers, the resulting node embeddings are denoted by $h_i$. The node embeddings parameterize
the discrete differential field over the graph. 

\subsection{Edge Decoder}

The node embeddings produced by the GNN encoder are then passed to the edge decoder, a multilayer perceptron (MLP) that predicts the raw transition score $r_{ij} = f_\theta(h_i \parallel h_j \parallel \mathbf{X})$ where $f_\theta(\cdot)$ is a multilayer perceptron. Although the continuous context $\mathbf{X}$ is already encoded implicitly within each node embedding $h_i$ through the GNN encoder. However, it is provided explicitly as input to the \ac{MLP}, as this provides the edge decoder with a direct, unattenuated copy of the context; this is analogous to a residual connection,  compensating for any dilution of context information that may occur during message passing.
To enforce antisymmetry of pairwise differences, the final edge prediction is defined as the average of the difference between $\omega_{ij}$ and $\omega_{ji}$
The resulting matrix $\Omega = [\omega_{ij}]$ is skew-symmetric, ensuring that predicted edge values form a valid
discrete $1$-form over the graph~\cite{desbrun2005discreteexteriorcalculus}.

\subsection{Training Objective}

The EFGN model is trained to approximate the
context-conditioned gradient field introduced in
Section~\ref{ssec:landscape}. Supervision is obtained from pairwise-energy differences computed from objective evaluations
of sampled combinations. The training objective combines
a regression loss that fits observed edge differentials with an additional regularization term that encourages the predicted
field to remain consistent with the gradient structure of the
energy landscape.

\subsubsection{Pairwise Differential Regression}

The primary supervision signal is derived from pairwise
energy differences between combinations~\cite{doi:10.1021/acs.jcim.1c00670}. For a fixed
context $\mathbf{X}$, the true differential field associates each
edge $(i,j)$ of the graph with $\omega_{\mathbf{X}}(i, j)$ Eq.~\eqref{omega}, which represents the change in the penalized objective when
transitioning from combination $\mathbf{Z}_i$ to $\mathbf{Z}_j$. These values provide direct observations of the local variation of the energy
landscape and therefore serve as the training targets.

Let $\hat{\omega}_\theta(i,j;\mathbf{X})$ denote the predicted edge differential produced by the model. Learning proceeds by
minimizing the discrepancy between predicted and observed
differences over the edges of the graph:

\[
L_{\text{diff}} =
\mathbb{E}_{\mathbf{X}}
\left[
\frac{1}{|E(G)|}
\sum_{(i,j)\in E(G)}
\ell\big(\hat{\omega}_\theta(i,j;\mathbf{X}), \omega_{\mathbf{X}}(i, j)\big)
\right],
\]

where $\ell(\cdot)$ denotes a regression loss such as the
mean squared error or Huber loss. In this paper, Huber loss is applied.

Each objective evaluation contributes supervision
for multiple combination pairs~\cite{doi:10.1021/acs.jcim.1c00670}. For a sampled combination
set, pairwise differences generate a dense set of training signals
that capture both the ordering and magnitude of energy changes
between combinations. As a result, the model learns a
structured approximation of how the objective varies across the
combinatorial space.

\subsubsection{Cycle Consistency Regularization}

While the pairwise differential regression loss supervises the predicted edge field with observed energy differences and serves as the core supervisory signal, it does not guarantee that the learned edge field corresponds to a valid gradient field of an underlying energy landscape. As discussed in Section~\ref{ssec:landscape}, the true gradient field induced by the energy function satisfies cycle consistency: the cumulative energy change along any closed loop of combinations must vanish. In practice, however, the model predicts an approximate edge field $\hat{\omega}_\theta(i,j;\mathbf{X})$, which may exhibit local inconsistencies due to approximation error, limited sampling of the combinatorial space, or modeling noise. These inconsistencies manifest as non-zero circulation of energy around cycles in the graph, indicating deviations from an integrable gradient structure.

To encourage the learned edge field to remain consistent with the underlying potential structure, we introduce a \textbf{cycle-consistency regularizer} that penalizes circulation around graph cycles. In complete graphs, the smallest closed loops correspond to triangular cycles. For a triangle $(i,j,k)$, the cycle residual is defined as $c(i,j,k;\mathbf{X})$ and the cycle consistency loss as $L_{\text{cycle}}$, both are formulated as:
\begin{equation*}
    c(i,j,k;\mathbf{X}) = \hat{\omega}_\theta(i,j;\mathbf{X})+\hat{\omega}_\theta(j,k;\mathbf{X})+\hat{\omega}_\theta(k,i;\mathbf{X})
\end{equation*}

\begin{equation*}
    L_{\text{cycle}} = \mathbb{E}_{\mathbf{X}}\left[\frac{1}{|\mathcal{T}|}\sum_{(i,j,k)\in\mathcal{T}}c(i,j,k;\mathbf{X})^2\right]
\end{equation*}

where $\mathcal{T}$ denotes the set of triangular cycles in the graph. In practice, the loss is computed over a randomly sampled subset of cycles at each training step, providing an unbiased stochastic approximation of the full cycle-consistency objective. Minimizing this loss suppresses local rotational (non-conservative) components in the predicted edge field. It encourages the learned field to lie close to the exact gradient subspace, from which a scalar energy representing a global ranking can be recovered.

\subsubsection{Overall Objective}

The final training objective combines the differential regression loss with the structural regularization terms introduced above.  The overall training objective is therefore defined as
\begin{equation}
L = L_{\text{diff}} + \lambda_{\text{cycle}} L_{\text{cycle}} 
\end{equation}
where $\lambda_{\text{cycle}}$ control the contribution of the cycle consistency term. 

\section{Optimization Framework}
\subsection{Overview of the GNN-ReCo Optimization Framework}

We adopt the GNN-ReCo optimization framework, which decomposes the design problem into continuous variables $\textbf{X}$ and discrete combination choices $\textbf{Z}$, with objective $f_E(\textbf{Z}, \textbf{X})$. The framework alternates between exploring the continuous design space and selecting optimal combinations conditioned on each candidate design. At each iteration, a population-based optimizer, specifically \ac{MDPSO}~\cite{chowdhury2013mixed}, generates candidate continuous vectors $\textbf{X}^{(t)}$. For each candidate, a neural model predicts the optimal combination.
\begin{equation}
\textbf{Z}^* = \arg\min_{\textbf{Z} \in \mathbb{Z}} f_E(\textbf{Z}, \textbf{X}^{(t)}),
\end{equation}
allowing the optimizer to focus on continuous search while delegating combinatorial selection to the model. In the present work, the GNN is replaced with the trained EFGN model, \textbf{GNN-NavCo}, which predicts pairwise energy differences between combinations, providing a structured representation of the local energy landscape. Figure~\ref{fig:Overall Framework of GNN-NavCo Optimization} shows the framework of the training of GNN-NavCo and the optimization with GNN-NavCo. Note that \ac{MDPSO} is selected for demonstration purposes, and the proposed framework is in theory compatible with other standard gradient-based or gradient-free constrained non-linear optimization solvers as well. The observable benefits of \textbf{GNN-NavCo} is however likely to have some dependency on the dynamics (principles) driving the search process in different algorithms. To provide initial illustration of this cross-algorithm applicability, we also apply \ac{GA} as the solver for a subset of the problems, as described in Section~\ref{ssec:ga}. The following subsection describes how this differential model is applied to candidate combination subgraphs during optimization.

\begin{figure*}[t] 
    \centering
    \includegraphics[width=\textwidth]{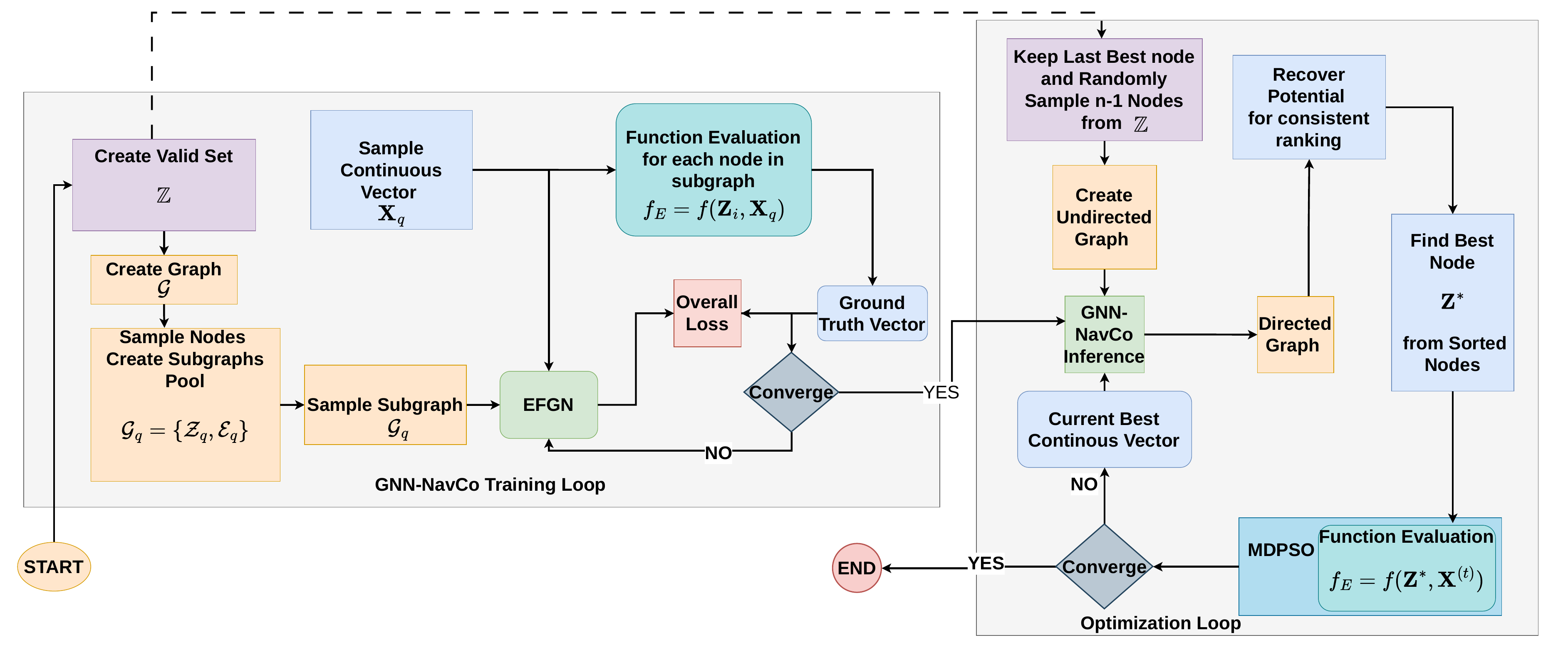}
    \caption{Overall Framework of GNN-NavCo Optimization}
    \label{fig:Overall Framework of GNN-NavCo Optimization}
\end{figure*}

\subsection{Inference on Candidate Subgraphs}

At each iteration, the continuous optimizer proposes a design vector $\textbf{X}^{(t)}$, which serves as the continuous context $\textbf{X}$ for GNN-NavCo. As Fig.~\ref{fig:gnn_navco} shows, a candidate subgraph is constructed by sampling a pool of combinations, with the current best combination $\mathbf{Z}^*$ included. The remaining nodes are sampled stochastically to ensure diversity and reduce sampling bias, consistent with the subgraph-based learning paradigm used in GNN-ReCo~\cite{gnn_reco}. The resulting input is an \textit{undirected fully connected graph} over sampled combinations and the continuous vector $\textbf{X}^{(t)}$. GNN-NavCo transforms this into a \textit{directed fully connected graph}, where, predictions form an antisymmetric matrix $Y \in \mathbb{R}^{N \times N}$ representing
the learned differential field over the candidate graph, where $N$ is the number of nodes in the graph. This can be interpreted as assigning a directional ``flow'' over the graph, indicating which transitions are favorable under the current context.

From an intuitive standpoint, if we anchor ourselves at the current combination $\mathbf{Z}^*$, the outgoing edges describe local directional preferences—i.e., which neighboring combinations appear better. However, this local view is inherently myopic: it does not account for how each candidate compares globally with all other combinations. To resolve this, we aggregate the predicted pairwise differences across the entire subgraph and recover scalar potentials for each node. This recovery step integrates the distributed edge information into a globally consistent ranking, enabling selection of the next combination. Importantly, because the graph is fully connected and includes $\textbf{Z}^*$, this global aggregation remains consistent with the local decision perspective at $\textbf{Z}^*$, while correcting for inconsistencies in the predicted field. As a result, the method effectively provides a \textit{globally informed discrete update direction}, giving the optimizer a gradient-like signal over the combinatorial space without violating its discrete structure. For a fully connected candidate graph this reduces to $\Phi_i$. The recovered potentials provide a scalar ranking of candidate combinations, and the combination with the minimum potential is selected. In practice, potential recovery reduces to a simple averaging operation.

\begin{equation*}
    \Phi_i = \frac{1}{N} \sum_{j=1}^{N} Y_{ij}
\end{equation*}

\begin{equation*}
    \textbf{Z}^* = \arg\min_i \Phi_i
\end{equation*} 

This procedure enables efficient evaluation of candidate combination subsets while preserving the relative structure of the energy landscape predicted by the differential model. In PyTorch, this can be implemented as $\Phi = \text{mean}(Y, \text{dim}=1)$, which aggregates the predicted pairwise energy differences for each combination.

\section{Evaluation: GNN-NavCo Optimization on Benchmark Problems}
We design a set of case studies to evaluate the GNN-NavCo framework on benchmark problems derived from MINLPLib~\cite{minlp}. The studies aim to: \textbf{1)} assess scalability with respect to the size of the combinatorial set; \textbf{2)} evaluate scalability as problem dimensionality increases; and \textbf{3)} examine applicability to both unconstrained and constrained problems.
Three minimization problems are selected from MINLPLib, including two unconstrained problems, \textit{cvxnonsep\_psig20} and \textit{cvxnonsep\_psig40}, and one constrained problem, \textit{cvxnonsep\_normcon40}. These are convex mixed-integer nonlinear programming problems. We construct a finite combination set by fixing the integer variables to predefined values, thereby transforming the original MINLP into an MCNLP formulation, and solve them with the \textbf{GNN-NavCo}-aided optimizer to compare with the original optimizer on performance metrics. The combination sets used in the paper are available on GitHub~\cite{githubrepo_navco}. The optimizer used here is \ac{MDPSO}, a state-of-the-art heuristic optimization algorithm~\cite{chowdhury2013mixed}. For both \ac{MDPSO} and GNN-MDPSO, the population size is set to 101, and the total number of iterations is set to 100. The optimization is set to automatically terminate if no improvement in feasibility or objective value is observed over 15 consecutive iterations. All the other parameters are set to the default.

\subsubsection{Unconstrained problem:}
We first use an unconstrained problem as the experimental setup, the \textit{cvxnonsep\_psig20} benchmark, which consists of 10 discrete and 10 continuous variables and imposes no constraints. The combinatorial set is represented by $\mathbb{Z}$, a candidate pool consists of $\mathbf{Z}$, comprising 101 combinations, each corresponding to a specific assignment of the discrete variables. To evaluate scalability, we further extend the experiments to larger combination pools containing 501 and 1001 candidates. Increasing the size of $\mathbb{Z}$ allows us to systematically study the robustness and stability of the proposed approach as the combinatorial search space expands, while also assessing its ability to learn a coherent representation of the underlying energy landscape under increasing complexity. We further extend the study to the \textit{cvxnonsep\_psig40} benchmark, which increases the problem dimensionality to 20 discrete and 20 continuous variables. This setting introduces a significantly larger and more complex combinatorial space. For consistency, the candidate pool is fixed to 101 combinations, and performance is evaluated under the same sampling and training protocol. This experiment isolates the effect of increased problem dimensionality while keeping the sampling budget constant. 

\subsubsection{Constrained problem:}
We then consider \textit{cvxnonsep\_normcon40}, which consists of 20 discrete and 20 continuous variables with one constraint. The candidate pool is fixed to 101 combinations, similar to the unconstrained setting. In this case, the presence of constraints introduces additional structure into the energy landscape through a penalty term.

\subsubsection{Data Sampling}
Based on the GNN-ReCo study, we sample continuous variables using Latin Hypercube Sampling (LHS) within their prescribed bounds, providing a space-filling design that ensures uniform coverage of the continuous domain. For each sampled continuous context $\mathbf{X}_q$, a candidate subgraph $\mathcal{G}_q$ is constructed by randomly selecting nodes from the global combination pool. This stochastic sampling ensures broad and unbiased coverage of the combinatorial space while maintaining computational efficiency. Given a sampled pair $(\mathcal{G}_q, \mathbf{X}_q)$, all combinations within the subgraph are evaluated under the same continuous context to obtain their corresponding scalar energy values. From these node-wise evaluations, pairwise energy differences are computed to define directed edge attributes, yielding a fully connected graph representation in which each edge encodes the relative energy difference between two combinations. 

\begin{figure*}[h!]
    \centering
    \includegraphics[width=\linewidth]{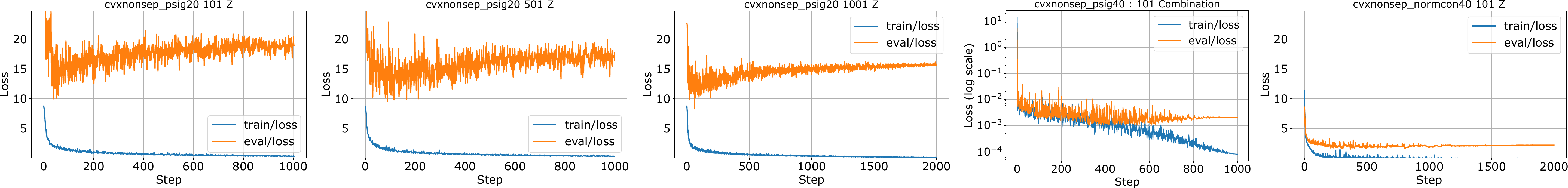}
    \caption{Training and validation loss of 5 EFGN Models for each of the benchmark problems. For \textit{cvxnonsep\_psig20}, EFGN is trained 3 times separately with 101, 501, and 1001 combinations. For \textit{cvxnonsep\_normcon40} and \textit{cvxnonsep\_psig40}, EFGN is trained with 101 combinations each. For \textit{cvxnonsep\_psig40}, the loss values were between 0-1, therefore plotted against log loss for resolution.}
    \label{fig:loss}
\end{figure*}

\section{Results and Discussions}
\subsection{GNN Training and Evaluation}
Figure~\ref{fig:loss} shows the training and validation loss of all the GNN-based models we used for the upcoming optimization case studies. For each case, 2,400 samples are used for training and 600 samples are used for validation. Training loss converges in all cases; the validation loss does keep oscillating in some of the cases. Hence, based on the validation loss history, we save the model checkpoint with the lowest validation loss, i.e., before overfitting occurs in each case.


\begin{figure*}
    \centering
    \includegraphics[width=0.7\linewidth]{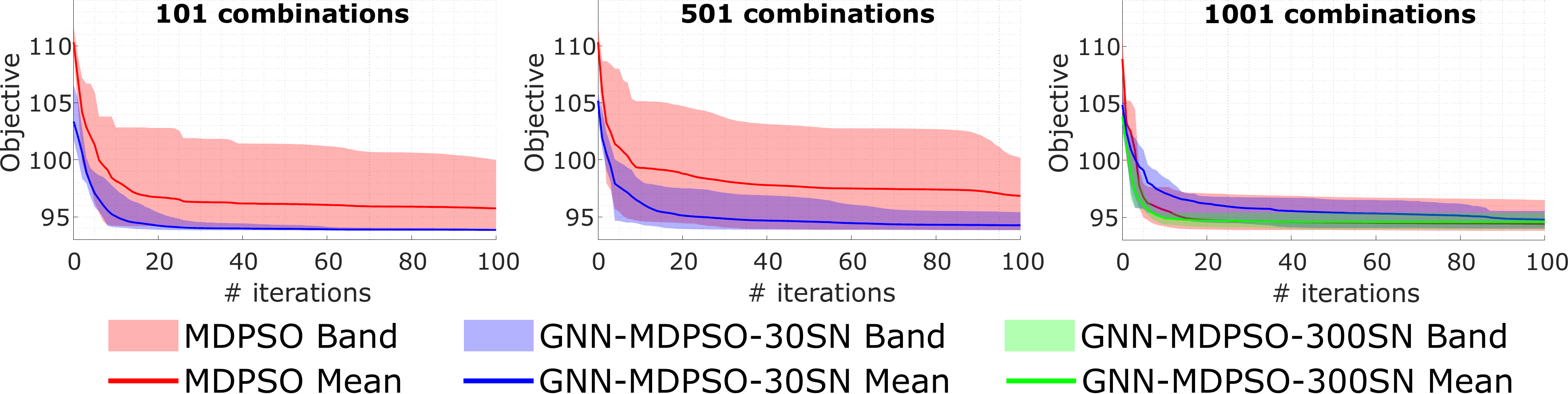}
    \caption{Objective (Energy) convergence history of MDPSO and GNN-MDPSO on the \textit{cvxnonsep\_psig20} with different numbers of valid combinations considered in the optimization. For the 1001-combinations optimization, GNN-NavCo trained on 300-node subgraphs is used, whereas the other two optimizations use GNN-NavCo trained on 30-node subgraphs.  }
    \label{fig:psig20}
\end{figure*}

\begin{figure}
    \centering
    \includegraphics[width=1.0\linewidth]{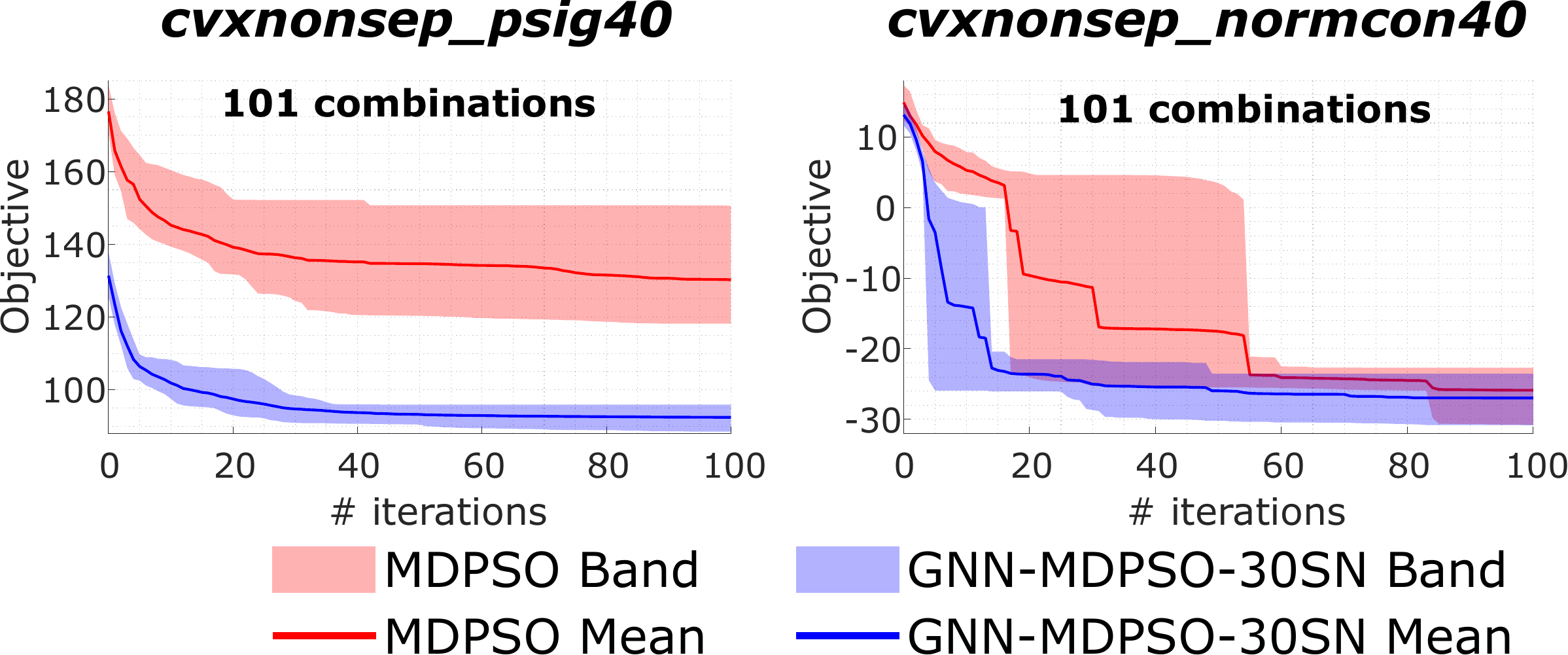}
    \caption{Objective (Energy) convergence history of MDPSO and GNN-MDPSO on the unconstrained problem \textit{cvxnonsep\_psig40} and constrained problem \textit{cvxnonsep\_normcon40}.}
    \label{fig:conv40}
\end{figure}

\begin{figure*}
    \centering
    \includegraphics[width=0.9\linewidth]{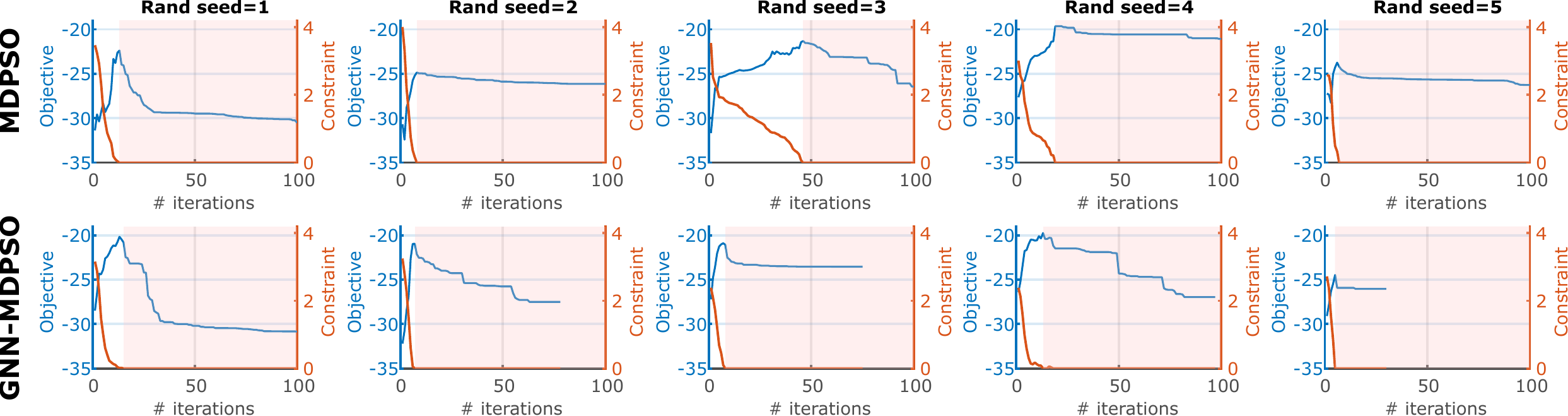}
    \caption{Objective and Constraint Convergence History of MDPSO with standard constraint formulation and GNN-MDPSO with penalty function (with constraint and objective recalculated) on the constrained problem \textit{cvxnonsep\_normcon40}. The pink shaded area represents the feasible region.}
    \label{fig:norm40_YY}
\end{figure*}

\subsection{Unconstrained benchmark problem}\label{ssec:uncons}
Figure~\ref{fig:psig20} shows the objective convergence histories of the unconstrained optimization case studies and the energy convergence history of the constrained optimization case studies. Table~\ref{tab:case_study_result} shows the mean optimized objectives of each case study, and the optimal method has been bolded. In most cases, the \textbf{GNN-NavCo}-aided \ac{MDPSO} (denoted as GNN-MDPSO) leads to better results compared with the original MDPSO.

\textbf{\textit{cvxnonsep\_psig20}}: The convergence history of the optimization on \textit{cvxnonsep\_psig20} is shown in Fig.~\ref{fig:psig20}. The shaded region represents the range (minimum to maximum) of the objective values over five independent runs with different initial populations and random seeds. Within each run, both the original \ac{MDPSO} and the GNN-MDPSO share the same initial population. However, their convergence histories begin from different starting points. This discrepancy arises because, in GNN-MDPSO, the initial population is first processed by \textbf{GNN-NavCo}, which recommends alternative combinations for function evaluation, thereby altering the effective starting point of the optimization process as expected. When the size of the combination set is 101, the GNN-MDPSO optimizes the objective to 93.855, which is only 0.044 worse than the true optimal (93.811). The GNN-MDPSO converges 12.93\% more than the original \ac{MDPSO}. Furthermore, the GNN-MDPSO is much more robust, leading to the convergence history with much less variation compared with the original \ac{MDPSO}. When the size of the combination set increases to 501, both methods show higher variances due to the increase in the combinatorial search space, but the GNN-MDPSO still holds a better converged objective and variance in the five runs. As the size further increases to 1,001, the variance of GNN-MDPSO remains relatively stable; however, its convergence rate becomes slower than that of the original \ac{MDPSO}. A plausible explanation is that the number of nodes sampled in the subgraph is insufficient to effectively explore the much larger combinatorial space. To investigate this, five additional experiments are conducted with an increased subgraph size of 300 nodes (denoted as GNN-MDPSO-300SN). As shown in Fig.~\ref{fig:psig20}, both convergence performance and variance are significantly improved with more nodes in the subgraphs.

\textbf{\textit{cvxnonsep\_psig40}}: After doubling the number of continuous and combinatorial variables, GNN-MDPSO maintains strong convergence performance, as the left plot in Fig.~\ref{fig:conv40} shows. The final optimized objective remains significantly lower than that obtained by the original \ac{MDPSO}, while exhibiting reduced variation across runs. These results indicate that GNN-MDPSO retains a high level of robustness as the dimensionality of the design space increases. This suggests that the proposed framework has scalability with respect to problem complexity and is capable of effectively handling higher-dimensional mixed combinatorial optimization problems without significant degradation in performance.



\begin{table}[h]
\footnotesize
    \centering
    \begin{threeparttable}
    \caption{Mean Optimized Objective of the Case Studies over 5 Runs}
    \label{tab:case_study_result}
    \begin{tabular}{llllll}
        \toprule
        Problem & Method & Error w.r.t. True Optimum  \\
        \midrule
        \multirow{2}{*}{ \makecell[l]{\textit{cvxnonsep\_psig20} \\ (101 Combination)}}
                               & MDPSO & 1.930    \\
                               & GNN-MDPSO-30SN &\textbf{ 0.044}   \\
        \hdashline
        \multirow{2}{*}{ \makecell[l]{\textit{cvxnonsep\_psig20} \\ (501 Combination)}}
                               & MDPSO & 3.029    \\
                               & GNN-MDPSO-30SN & \textbf{0.446}  \\
        \hdashline  
        \multirow{3}{*}{ \makecell[l]{\textit{cvxnonsep\_psig20} \\ (1001 Combination)}}
                               & MDPSO & \textbf{0.595}    \\
                               & GNN-MDPSO-30SN & 0.984   \\
                               & GNN-MDPSO-300SN & 0.777 \\
        \hdashline        
        \multirow{2}{*}{ \makecell[l]{\textit{cvxnonsep\_psig40} \\ (101 Combination)}}
                               & MDPSO & 48.398    \\
                               & GNN-MDPSO-30SN & \textbf{9.918}   \\
        \hdashline
        \multirow{3}{*}{ \makecell[l]{\textit{normcon40} \\ (101 Combination)}}
                               & MDPSO & 6.75    \\
                               & MDPSO-Cons* & 6.631 \\
                               & GNN-MDPSO-30SN & \textbf{5.651}   \\
        \bottomrule 
    \end{tabular}
    \begin{tablenotes}
        \item[*] This MDPSO is run with the standard constraint formulation without utilizing the penalty function. Error w.r.t. true optimum is reported in terms difference in the objective function value. 
    \end{tablenotes}
    \end{threeparttable}
\end{table}


\subsection{Constrained benchmark problem}
\textbf{\textit{cvxnonsep\_normcon40}}: After switching the problem to constrained optimization problem, Fig.~\ref{fig:conv40} shows the convergence history of the two methods. It can be observed that the GNN-MDPSO is still robust and converging much faster, and leads to the mean objective being more optimal than the original MDPSO. Note that, in this figure, the MDPSO also runs with the penalty function.

Figure~\ref{fig:norm40_YY} shows the constraint violations and the objective convergence history of the MDPSO with standard constraint formulation and the GNN-MDPSO with penalty functions, which can provide a perspective on whether the order-preserving search combined with the penalty function can efficiently lead the optimizer to the feasible region. It shows that the GNN-MDPSO enters the feasible region (the shaded section) faster than the original MDPSO, proving that the order-preserving search works well in the constrained problem as well. Also, 3 out of 5 runs the GNN-MDPSO achieved more optimal objectives, which indicates the potential of using this GNN-aided optimization concept on an unconstrained solver in the future as well.


\begin{figure*}
    \centering
    \includegraphics[width=1.0\linewidth]{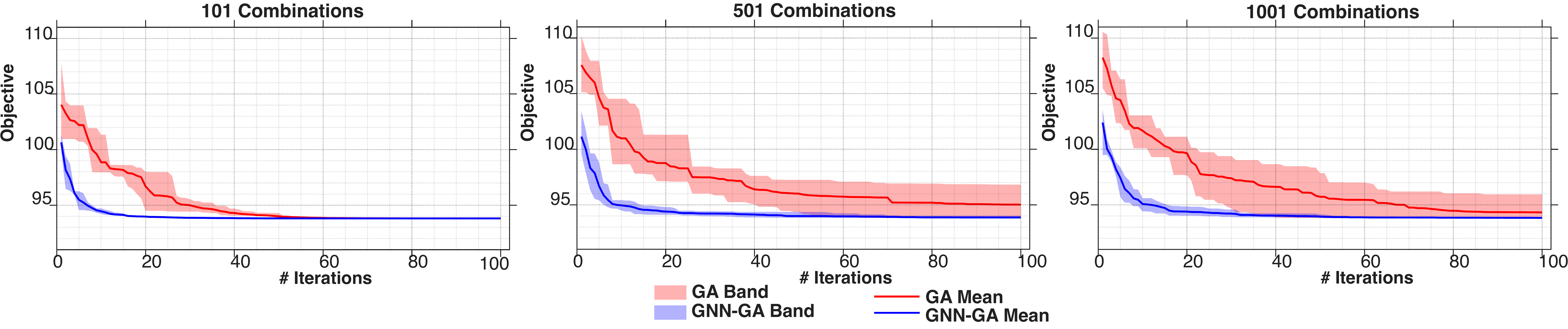}
    \caption{Objective (Energy) convergence history of GA and GNN-GA on the unconstrained problem \textit{cvxnonsep\_psig20}}
    \label{fig:ga_result}
\end{figure*}

\subsection{GNN-NavCo-aided GA on unconstrained problems}\label{ssec:ga}
To demonstrate that the proposed framework can work with optimizers other than the 
\ac{MDPSO}, the optimizer is replaced with \ac{GA} implemented with MATLAB \texttt{ga()}, and the framework is tested on the \textit{cvxnonsep\_psig20} problem with the identical three combination sets explained in Section~\ref{ssec:uncons}. The population size and the max generation are set to 100. All the other hyperparameters are the same as the default settings. The \ac{GA} has been run five times with and without the aid of the GNN-NavCo as comparisons to show the benefits of the GNN-NavCo. In each case, the corresponding GNN-NavCo in Section~\ref{ssec:uncons} is used, which has been trained with 2,400 samples and validated with 600 samples, and the subgraph size is 30-node.

Figure~\ref{fig:ga_result} shows the convergence histories of the objective function values. In the optimization experiments with 101, 501, and 1,001 combinations, it can be observed that the incorporation of GNN-NavCo significantly accelerates convergence compared to the GA without the aid of GNN. For the 101-combination case, the GNN-NavCo-aided GA reaches convergence approximately 30 generations earlier, corresponding to about 3,000 fewer function evaluations.
For the larger combination spaces (501 and 1,001), the baseline GA not only converges more slowly but also yields noticeably worse mean converged objective values. In contrast, the GNN-NavCo-aided GA converges approximately 40 and 50 generations faster, respectively, while achieving superior final performance.

These results indicate that as the size of the combinatorial space increases, the benefit of GNN-NavCo becomes more pronounced, suggesting its effectiveness in guiding the search through increasingly complex design spaces. Furthermore, the convergence behavior of the GNN-NavCo-aided GA is significantly more stable, exhibiting reduced variance across different random initial populations and improved robustness compared to the GA without the aid of GNN.

The significantly larger improvement observed in the GA-based framework, compared to the MDPSO-based framework, suggests that the GNN-NavCo primarily compensates for limitations associated with the integer-based representation of combinatorial variables. In the standard GA formulation, crossover and mutation operations are applied directly to integer indices that do not necessarily preserve meaningful similarity relationships among combinations. As a result, the search process may become inefficient in highly nonlinear combinatorial spaces. In contrast, MDPSO inherently maintains smoother search dynamics in mixed-variable optimization problems, reducing the relative benefit provided by the GNN-NavCo. This indicates that the primary advantage of the GNN mechanism lies in its ability to provide structure-aware guidance for combinatorial exploration.

\subsection{Computing Cost Comparison}
It is observed that in the most above cases, the number of iterations that GNN-MDPSO and GNN-GA spent to reach convergence is approximate 20 iterations (except in \textit{cvxnonsep\_psig20} with 1001 combinations for MDPSO) fewer than the original Baseline. Since the population size of both methods is 101, this leads to approximately 2,020 fewer function evaluations for the GNN-Aided method to converge. Considering in all the above case studies, the training sets of the GNNs are generated with 2,400 function evaluations, the cost of the GNN-Aided method is about the same as the cost of the original MDPSO to converge. Also, over repeated optimization runs, the effective computational cost of the GNN-assisted approach becomes significantly lower than that of the baseline method, while also providing faster and more reliable convergence characteristics. When the convergence band is interpreted probabilistically, then for cases such as Fig.~\ref{fig:psig20}, where the convergence bands of MDPSO and GNN-MDPSO overlap, the GNN-assisted method still demonstrates a higher likelihood of reaching a near-optimal solution within a fixed iteration budget (e.g., at 40 iterations). In other words, although both methods may exhibit similar convergence behavior, the distribution of outcomes indicates that GNN-MDPSO and GNN-GA converge more reliably toward lower objective values. For the remaining case studies in Fig.~\ref{fig:conv40} and \ref{fig:ga_result}, the advantage becomes more pronounced for the GNN-NavCo method as it consistently remains below baseline across nearly all iterations. This indicates a higher probability of achieving better objective values throughout the optimization process when using the GNN-assisted methods.


\section{Conclusion}

This paper developed an approach to abstract combinatorial spaces in mixed-combinatorial non-linear programming (MCNLP) problems, in a manner that allows scalable and interpretable recommendation of the combinations to go with every candidate design in an optimization process that searches over the remaining non-combinatorial (e.g., continuous variables). In doing so, it seeks to fill the gap in the availability of general-purpose solvers or algorithms for MCNLP problems. Our model formulation, called GNN-NavCo, uses pairwise differences of energy values between combinations for a given context. This formulation can be interpreted as a mathematical object that resembles a graph with a superimposed field on top of it for each of the continuous contexts that may exist. Its use as an order-preserving combination recommender is demonstrated using a recently proposed decomposed solution framework for MCNLPs.  is applied to a suite of analytical nonlinear benchmark problems with up to 20 continuous variables and 20 combinatorial features, and 100's of feasible combinations. The training process was found to be effective in providing stable learning of the GNN-based recommender. When used in optimization with PSO, it was found to usually provide better optimum solutions (except in a few cases) compared to a baseline optimization that uses indexification of the combinations. This is with similar overall function evaluations taken into consideration, where we account for the additional investment required to train the GNN in our approach. From solely the optimization history, with GNN-NavCo, convergence was found to be substantially faster than that with the baseline. These benefits were found to be even more pronounced when GNN-NavCo is applied with GA as the optimizer (versus a baseline GA using indexification). Significantly lower variance in both the optimization history and final values is also observed compared to the baseline. It was also found that increasing the size of the sampled sub-graph (used for abstraction and then recommendation) with an increase in the number of combinations in the problems is helpful. However, more work is needed in the future to systematically identify sub-graph sizes (or a smarter sampling technique) to offset any comparative performance loss (as currently observed) compared to baselines for problems with a larger number of combinations.

While in its current form, GNN-NavCo is used in conjunction with PSO and \ac{GA}, the underlying recommendation process and model form is in principle not restricted to these optimization algorithms alone. So, an immediate direction of future work will be to use GNN-NavCo in conjunction with other standard gradient-based algorithms (e.g., sequential quadratic programming) and other gradient-free algorithms that serve as general-purpose NLP solvers. In addition, there remains an opportunity to explore better search mechanisms over the directed graph (given by GNN-NavCo) with probabilistic guarantees on order-preserving recommendations, informed by estimates of the GNN model uncertainty. Along with these improvements, applying GNN-NavCo to practical engineering design problems that present as complex MCNLPs would provide further insights into its benefits under realistic computational resource constraints.


\section*{Acknowledgments}
This work is supported under the CMMI Award numbered 2048020 from the National Science Foundation (NSF). The authors' opinions, findings, and conclusions or recommendations expressed in this material do not necessarily reflect the views of the National Science Foundation.

\bibliographystyle{IEEEtran}
\bibliography{sample} 

\end{document}